# Automated Prediction of Temporal Relations

Amol S Patwardhan, Jacob Badeaux, Siavash, Gerald M Knapp

## Abstract

a) Background: There has been growing research interest in automated answering of questions or generation of summary of free form text such as news article. In order to implement this task, the computer should be able to identify the sequence of events, duration of events, time at which event occurred and the relationship type between event pairs, time pairs or event-time pairs.

b) Specific Problem: It is important to accurately identify the relationship type between combinations of event and time before the temporal ordering of events can be defined. The machine learning approach taken in Mani et. al (2006) provides an accuracy of only 62.5 on the baseline data from TimeBank. The researchers used maximum entropy classifier in their methodology. TimeML uses the TLINK annotation to tag a relationship type between events and time. The time complexity is quadratic when it comes to tagging documents with TLINK using human annotation.

c) This research proposes using decision tree and parsing to improve the relationship type tagging. This research attempts to solve the gaps in human annotation by automating the task of relationship type tagging in an attempt to improve the accuracy of event and time relationship in annotated documents.

d) Scope information: The documents from the domain of news will be used. The tagging will be performed within the same document and not across documents. The relationship types will be identified only for a pair of event and time and not a chain of events. The research focuses on documents tagged using the TimeML specification which contains tags such as EVENT, TLINK, and TIMEX. Each tag has attributes such as identifier, relation, POS, time etc.

## Literature Review

Mani et. al (2006) used a maximum entropy classifier in their research. They assigned each pair of events one of 6 relations (before, after, simultaneous, includes, begins, ends) from an augmented Timebank corpus. The features included tense, aspect, modality, polarity and event class and were hand tagged. Lapata and Lascarides (2006) used supervised learning and developed an inter-sentential event classifier. A key time word such as before and after was used as trigger to indicate event and the sentences were saved with the tags to build a corpus. Syntax and clausal ordering features were used to train the classifier. Boguraev and Ando (2005) employed machine learning on relationship types between tasks instead of pair of event and time. Modi and Titov (2014) developed a classifier to identify the sequence of events in the text. In this research the events were represented as a combination of verbs and their arguments (subject, object etc.).

| Relation types | Our criterion | TimeML | Allen |
|---|---|---|---|
| EVENT 1<br>EVENT 2 | AFTER | AFTER | after |
| EVENT 1<br>EVENT 2 | | IAFTER | met-by |
| EVENT 1<br>EVENT 2 | OVERLAPPED−BY | | overlapped-by |
| EVENT 1<br>EVENT 2 | | ENDS | finishes |
| EVENT 1<br>EVENT 2 | DURING | DURING/IS_INCLUDED | during |
| EVENT 1<br>EVENT 2 | BEGUN_BY | BEGUN_BY | started-by |
| EVENT 1<br>EVENT 2 | SIMULTANEOUS | SIMULTANEOUS/IDENTITY | equal |
| EVENT 1<br>EVENT 2 | INCLUDES | INCLUDES/DURING_INV | contains |
| EVENT 1<br>EVENT 2 | ENDED_BY | ENDED_BY | finished-by |
| EVENT 1<br>EVENT 2 | OVERLAPS | | overlaps |
| EVENT 1<br>EVENT 2 | | BEGINS | starts |
| EVENT 1<br>EVENT 2 | BEFORE | IBEFORE | meets |
| EVENT 1<br>EVENT 2 | | BEFORE | before |

Fig. 1. Relation types between event tags. Cheng et. Al (2008).

## Methodology

a) The approach proposed in this research does not require training since it is using decision tree based classification and parsing. This research will use the following tagged corpus: Causal TimeBank (183 documents), AQUAINT (73 documents).

b) The inputs for the classification problem will be attributes from event tag, co-referential entities, POS tag, linear ordering of events and prepositional phrases and distance between events in terms of sentences, event class, aspect (adverb/ preposition), modality, tense and negation, timex3 time features.

c) The algorithm will be using the above features to identify a subset of relationship types shown in the Fig 1. Thus the algorithm will assign a class such as BEFORE, AFTER, SIMULTANEOUS to the event-event pair or event-timex pair based on the similarity in features.

d) Resulting semantic representation will contain a document that has TLINK tags establishing the relationship type between event and time tags from the document.

e) The relationship type identification can be seen as classification tasks, where a given temporal links is assigned a relation type from the set BEFORE, AFTER, or SIMULTANEOUS. Precision and recall over these relation types are used as evaluation metrics. The measurements will also be compared with the results obtained from supervised learning from Mani et. al (2006).

f) We will also manually annotate a sample set of 5 news articles related to sports, politics and international news.

We divided the relationship type objective into following two tasks.
Task 1) Identify the events in a document using words that are verbs in past tense and ending in ‘ed’. Identify the events in a document using the words that are verbs in present tense and ending in ‘ing’. Identify the events in a document using the words that are preceded by the work ‘will’. For this we developed a parser that adds an event tag conforming to the TimeML specification for each of the identified word.

Task 2) Categorize the pairs of events in the tagged document into three classes with label BEFORE, AFTER or SIMULTANEOUS by comparing the location and tense of the event. Create a TLINK tag conforming to the TimeML specification for each of the identified relationship type.

Task 3) Manually annotate the document and measure the recall and precision with the results obtained from decision tree based classification method used in task 2.

## Implementation

The user interface for the temporal relationship type detection program contains a file browser which can be used to load a document with news, sports, stories or any free form text. The detect events in text button tags the document with the EVENT tags for events occurring in the past using verbs in past tense. The detect timex in text button tags the document with the TIMEX3 tags for words indicate an instance of time. The ‘create tlink tags’ button uses the annotated text and performs classification using a decision tree and key word parsing on the document to identify before, after and simultaneous event-time relationship types and adds the tlinks to the document with the corresponding relationship type label. The compare manual annotation button compares the tlink tags generated by the system with the manually annotated tlink tags and calculates the precision and recall rate.

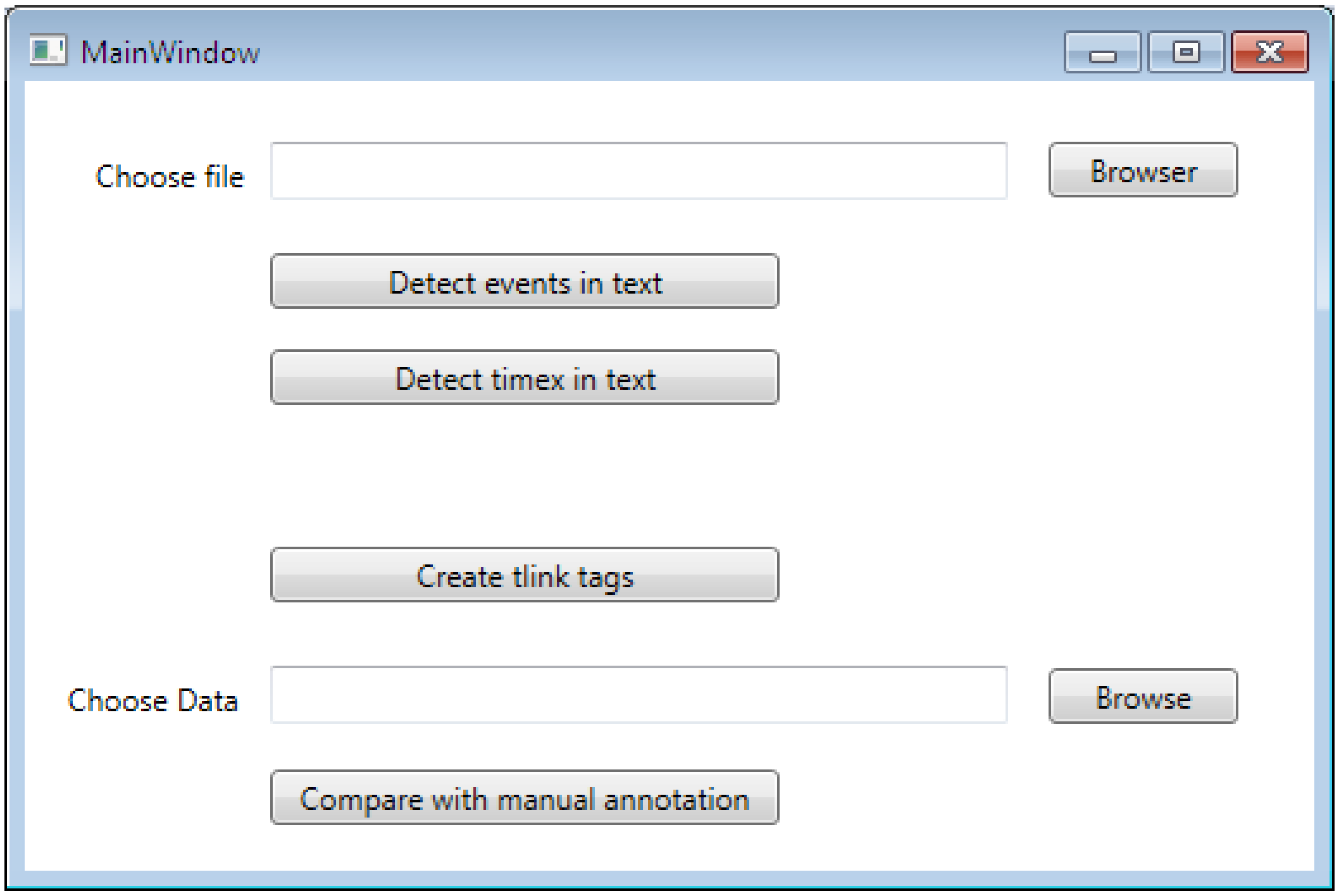

For the purpose of detecting the EVENT tags, we will use a set of action verbs which are commonly found based on the list obtained from cvisual.com as shown in Fig 2. The parser will compare the tokens to identify the verbs from the list and tag it as an EVENT.

For the purpose of establishing the relation type of BEFORE we will use the following list of key words: ahead, back, previously, since, sooner, afore, aforetime, ante, antecedently, anteriorly, before, present, ere, fore, former, formerly, forward, gone, gone by, heretofore, in advance, in days of yore, in front, in old days, in the past, past, precedent, previous, up to now.

2 words before and after the chosen action verb will be used to establish the relationship of the event tag.

For the purpose of identifying the TIMEX tags, the list of months (Jan, Feb, Mar, etc.), days of the weeks (Mon, Tue, Wed, etc.), words indicating time such as min, hours, evening, morning will be used.

If there are n events and tags, then there are n x n pairs possible. When a relation type down not exists, the TLINK tag will not be added to the document.

1) Example of parsing and decision making for SIMULTANEOUS relation type:

Sam ate an apple while he was doing homework.

Here ate and doing homework are the EVENTS

While is the keyword used to determine that the event occurred simultaneously.

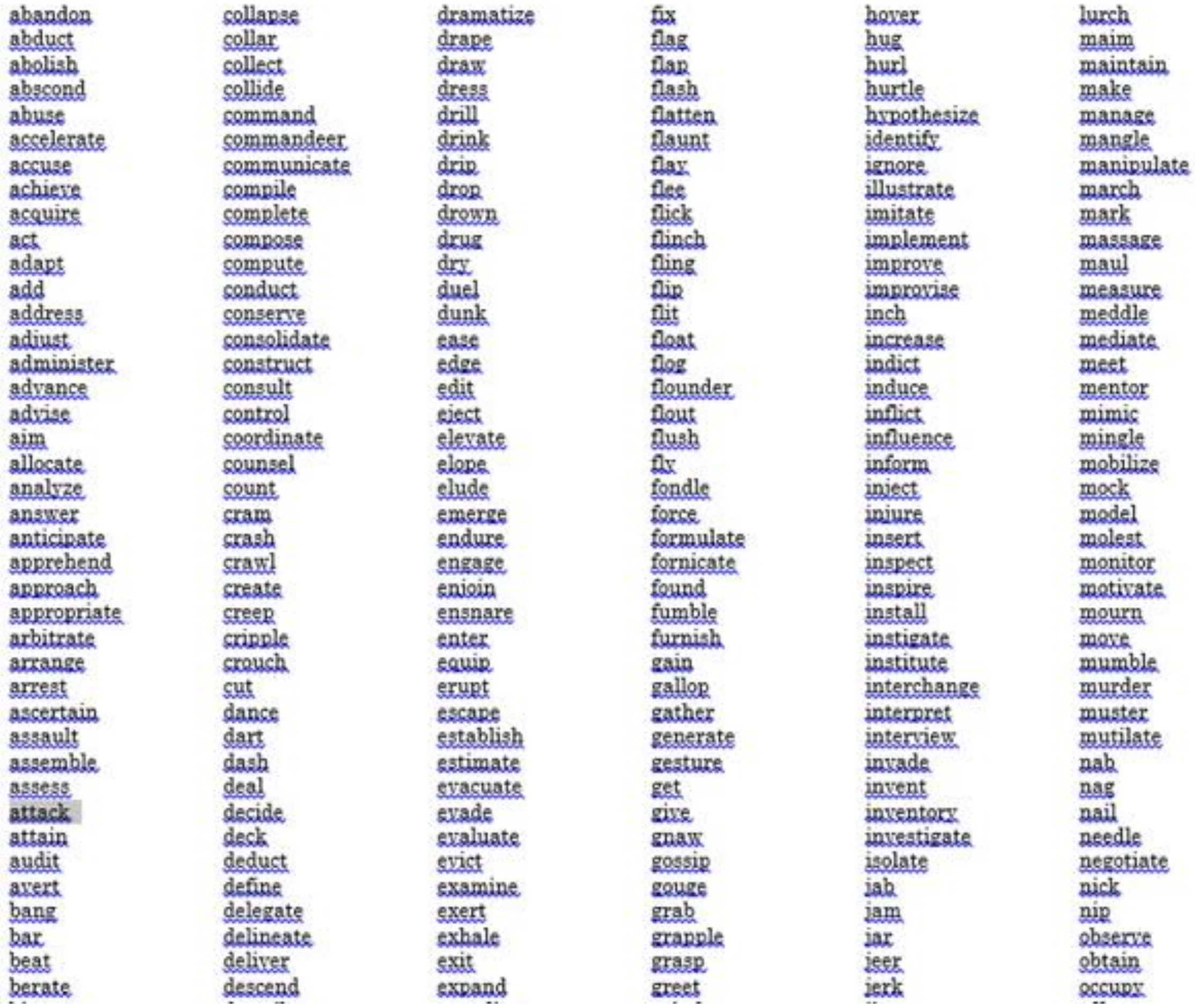

| | | | | | |
|---|---|---|---|---|---|
| abandon | collapse | dramatize | fix | hover | lurch |
| abduct | collar | drape | flag | hug | maim |
| abolish | collect | draw | flap | hurl | maintain |
| abscond | collide | dress | flash | hurtle | make |
| abuse | command | drill | flatten | hypothesize | manage |
| accelerate | commandeer | drink | flaunt | identify | mangle |
| accuse | communicate | drip | flay | ignore | manipulate |
| achieve | compile | drop | flee | illustrate | march |
| acquire | complete | drown | flick | imitate | mark |
| act | compose | drug | flinch | implement | massage |
| adapt | compute | dry | fling | improve | maul |
| add | conduct | duel | flip | improvise | measure |
| address | conserve | dunk | flit | inch | meddle |
| adjust | consolidate | ease | float | increase | mediate |
| administer | construct | edge | flog | indict | meet |
| advance | consult | edit | flounder | induce | mentor |
| advise | control | eject | flout | inflict | mimic |
| aim | coordinate | elevate | flush | influence | mingle |
| allocate | counsel | elope | fly | inform | mobilize |
| analyze | count | elude | fondle | inject | mock |
| answer | cram | emerge | force | injure | model |
| anticipate | crash | endure | formulate | insert | molest |
| apprehend | crawl | engage | fornicate | inspect | monitor |
| approach | create | enjoin | found | inspire | motivate |
| appropriate | creep | ensnare | fumble | install | mourn |
| arbitrate | cripple | enter | furnish | instigate | move |
| arrange | crouch | equip | gain | institute | mumble |
| arrest | cut | erupt | gallop | interchange | murder |
| ascertain | dance | escape | gather | interpret | muster |
| assault | dart | establish | generate | interview | mutilate |
| assemble | dash | estimate | gesture | invade | nab |
| assess | deal | evacuate | get | invent | nag |
| attack | decide | evade | give | inventory | nail |
| attain | deck | evaluate | gnaw | investigate | needle |
| audit | deduct | evict | gossip | isolate | negotiate |
| avert | define | examine | gouge | jab | nick |
| bang | delegate | exert | grab | jam | nip |
| bar | delineate | exhale | grapple | jar | observe |
| beat | deliver | exit | grasp | jeer | obtain |
| berate | descend | expand | greet | jerk | occupy |

Fig. 2. Sample list of action verbs used in the decision making process (www.cvisuals.com)

2) Example of parsing and decision making for BEFORE relation type:

Sam ate an apple and went to school.

Here ate and went are the EVENTS.

And is the keyword used to determine that the event ate occurred before event went.

In this case the AFTER relation type is implicit, where event went occurred after event ate.

The parser and decision code in this research will use such keywords in the neighborhood of the action verbs to determine the relation type between the event pair.

## Results

The system tagged a total of 25 documents on news articles from various fields such as politics, sports, and world news. Each automatically tagged document was compared with manually tagged temporal information. The results of the automated tagging system are described as follows.

| Tag | Manual tag count | Automated tag count |
|---|---|---|
| Event Tag | 2144 | 1992 |
| Timex Tag | 983 | 809 |
| TLINK (BEFORE) | 102 | 594 |
| TLINK (AFTER) | 578 | 731 |
| TLINK (SIMULTANEOUS) | 1172 | 852 |

The first column contains the type of tag such as event, timex, tlink and its sub types. The second column shows the count/ frequency of each tag when annotated manually. The third column shows the count/ frequency of each tag when generated by the automation code.

## Analysis

The performance was based on how well the automated tagging detected the tags in the news articles. The results of tagging the event tag indicate that the manual tagging performed better. The results of tagging the timex tag also indicate that the manual tagging performed better. The reason for the lower recall rate is that our proposed methodology only considered a subset of the verbs and as a result, the human annotator was able to outperform the proposed methodology. But the number of tags generated by the detector code can be improved using a more comprehensive list of action words.

The results for tagging the TLINK (BEFORE, AFTER) were better in case of automated tagging as compared to manual annotation. This is because the manual tagging can get cumbersome and the human annotator is not able to do a good job establishing relation types between events and times tags. Additionally, the methodology used parsing the neighboring words on both sides of the token and was computationally faster in calculating the relation type between the events as compared to human annotator who took time to annotate this information.

## Conclusion

The manual annotation is better in tagging the events and the time related information but the quantity and quality of the tags degrade when it comes to annotating the temporal relation type between events in a document. Based on the results the automated tagging certainly improved the recall rate on the TLINK tags which are important to store the temporal information between events and timex tags. The results indicate that automated temporal information tagging is certainly feasible task. For future scope we would focus on improving the precision of the detected relation type.